\pdfoutput=1

\documentclass[11pt]{article}

\usepackage[final]{acl}

\usepackage{times}
\usepackage{latexsym}

\usepackage[T1]{fontenc}

\usepackage[utf8]{inputenc}

\usepackage{microtype}

\usepackage{inconsolata}

\usepackage{graphicx}

\usepackage{amssymb}
\usepackage{bm}
\usepackage{lscape}
\usepackage{mathtools}

\title{Embedded Topic Models Enhanced by Wikification}

\author{Takashi Shibuya \quad Takehito Utsuro \\
  Degree Programs in Systems and Information Engineering,\\
  Graduate School of Science and Technology, University of Tsukuba\\
  \texttt{s2430171@\_u.tsukuba.ac.jp} \quad \texttt{utsuro@\_iit.tsukuba.ac.jp}}

\begin{document}
\maketitle
\begin{abstract}
Topic modeling analyzes a collection of documents to learn meaningful patterns of words.
However, previous topic models consider only the spelling of words and do not take into consideration the homography of words.
In this study, we incorporate the Wikipedia knowledge into a neural topic model to make it aware of named entities.
We evaluate our method on two datasets, 1) news articles of \textit{New York Times} and 2) the AIDA-CoNLL dataset.
Our experiments show that our method improves the performance of neural topic models in generalizability.
Moreover, we analyze frequent terms in each topic and the temporal dependencies between topics to demonstrate that our entity-aware topic models can capture the time-series development of topics well.
\end{abstract}

\section{Introduction}
\label{sec:intro}

Probabilistic topic models such as latent Dirichlet allocation (LDA)~\cite{10.5555/944919.944937} and embedded topic model (ETM)~\cite{dieng-etal-2020-topic} have been utilized for analyzing a collection of documents and discovering the underlying semantic structure.
Such topic models have also been extended to dynamic topic models~\cite{10.1145/1143844.1143859,hida-etal-2018-dynamic,dieng2019dynamicembeddedtopicmodel,Cvejoski_Sánchez_Ojeda_2023}, which can capture the chronological transition of topics, motivated by the fact that documents (such as magazines, academic journals, news articles, and social media content) feature trends and themes that change with time.

However, previous (dynamic) topic models consider only the spelling of words and do not take into consideration the homography of words such as ``\textit{apple}'' and ``\textit{amazon}''.
We hypothesize that this unawareness of the word homography harms the performance of topic models because one meaning of a word will tend to be used in some specific topics but another meaning of the same spelled word will appear in other topics more frequently.
For instance, the entity ``Amazon.com'' will tend to appear in business news or technology articles, whereas documents about the environment will discuss the entity ``Amazon rainforest'' more often than ``Amazon.com''.
Although the word ``\textit{Amazon}'' can thus refer to a different entity depending on a context, existing topic models are not aware of such homography of the word ``\textit{Amazon}'' and regard the word as unique.

To address the above issue, we propose a method of analyzing a collection of documents based on entity knowledge on Wikipedia.
Our proposed method relies on two technologies: 1) entity linking (wikification) and 2) entity embedding (Wikipedia2Vec~\cite{yamada-etal-2020-wikipedia2vec}).
Entity linking (wikification) is a natural language processing technique that assigns an entity mention in a document to a specific entity in a target knowledge base (Wikipedia).
For example, an entity linker can recognize which a word ``apple'' in a document means, ``Apple Inc.'', ``Big Apple'', or another.
We adopt entity linking as a preprocessing of topic modeling.
Next, we incorporate entity embeddings (vector representations of entities in a knowledge base) into a neural topic model according to the result of the entity linking.
Previous neural topic models utilize only conventional word embeddings, which are unaware of the homography of words.
On the other hand, our proposed method uses not only word embeddings but also entity embeddings, which enables neural topic models to distinguish between multiple entities that share their spelling.
We hypothesize that our entity-aware method improves the performance of neural topic models.
We empirically show the effectiveness of our method on two datasets: 1) a collection of news articles of \textit{New York Times} published between 1996 and 2020 and 2) the AIDA-CoNLL dataset~\cite{hoffart-etal-2011-robust}.
We adopt two topic models, ETM and dynamic ETM~\cite{dieng2019dynamicembeddedtopicmodel}, as baselines and quantitatively show that entity linking improves the performance of neural topic models.
Furthermore, we demonstrate that topics and their temporal change extracted by trained dynamic topic models are reasonable by manually analyzing frequent terms of each topic.
We summarize our contributions as follows:
\begin{itemize}
    \item We propose a method to make neural topic models aware of named entities. Our method utilizes entity linking (wikification) as preprocessing and incorporates entity embeddings (Wikipedia2Vec) into neural topic models.
    \item We quantitatively demonstrate that our proposed method improves the performance of neural topic models on a dataset containing many homographic words such as ``apple''.
    \item We manually analyze topics extracted by trained topic models and verify that our proposed method brings high interpretability because frequent terms in each topic are expressed with Wikipedia entries.
    \item We also show that our method does not harm the performance even on a dataset that does not include many homographic words (if entity linking is accurate enough).
\end{itemize}

\section{Related Work}
\label{sec:related_work}

\subsection{Neural Topic Models}

Our method builds on a combination of topic models and word embeddings, following a surge of previous methods that leverage word embeddings to improve the performance of probabilistic topic models.
Some methods incorporate word similarity into the topic model~\cite{NIPS2010_db85e259,xie-etal-2015-incorporating,8215536}.
Other methods combine LDA with word embeddings by first converting the discrete
text into continuous observations of embeddings~\cite{das-etal-2015-gaussian,batmanghelich-etal-2016-nonparametric,7837989,ijcai2017p588}.
Another line of research improves topic modeling inference utilizing deep neural networks~\cite{pmlr-v70-cong17a,zhang2018whai,card-etal-2018-neural}.
These methods reduce the dimension of the text data through amortized inference and the variational auto-encoder~\cite{pmlr-v32-kingma14}.
Finally, \citet{dieng-etal-2020-topic} proposed the embedded topic model (ETM) that makes use of word embeddings and uses amortization in its inference procedure.

\subsection{Dynamic Topic Models}

The seminal work of \citet{10.1145/1143844.1143859} introduced dynamic latent Dirichlet allocation (D-LDA), which uses a state space model on the parameters of a topic distribution, thus allowing the distribution parameters to change with time.
\citet{dieng2019dynamicembeddedtopicmodel} proposed an extension of D-LDA, dynamic embedded topic model (D-ETM), that better fits the distribution of words via the use of distributed representations for both the words and the topics.
Furthermore, \citet{miyamoto-etal-2023-dynamic} introduced the self-attention mechanism into the neural network used in amortized variational inference.

\subsection{Entity Embeddings}

Entity embeddings have been studied mainly in the context of named entity disambiguation (NED).
\citet{Bordes_Weston_Collobert_Bengio_2011,NIPS2013_b337e84d,Lin_Liu_Sun_Liu_Zhu_2015} focus on knowledge graph
embeddings and propose vector representations of entities to primarily address the knowledge base (KB) link prediction task.
\citet{wang-etal-2014-knowledge} proposed the joint modeling of the embedding of words and entities and revealed that such joint modeling improves performance in several entity-related tasks including the link prediction task.
\citet{yaghoobzadeh-schutze-2015-corpus} built embeddings of words and entities on a corpus with annotated entities using the skip-gram model to address the entity typing task.
Finally, \citet{yamada-etal-2016-joint} proposed an embedding method
that consists of three models: 1) the conventional skip-gram model that learns to predict neighboring words given the target word in text corpora, 2) the anchor context model that learns to predict neighboring words given the target entity using anchors and their context words in the KB, and 3) the KB graph model that learns to estimate neighboring entities given the target entity in the link graph of the KB.
To the best of our knowledge, our study is the first attempt to incorporate entity embeddings into embedded topic models.

\subsection{Topic Models with Wikipedia}

There have been several works where topic models are applied to Wikipedia.
Most such studies worked on cross-lingual topic modeling by harnessing Wikipedia's cross-linguality~\cite{10.1145/1526709.1526904,10.5555/1795114.1795124,10.5555/2540128.2540447,hao-paul-2018-learning,10.1145/3442381.3449805}.
In Wikipedia, each article describes a concept, and each concept is usually described in multiple languages.
They proposed formulations of cross-lingual topic models and verified the efficacy of their proposed topic models trained on Wikipedia articles and links.
Aside from the above studies, \citet{10.1145/3366424.3383567} applied topic models to Wikipedia for analyzing popular topics in different language editions.
In contrast to these works, our method utilizes Wikipedia entities identified by entity linking to make embedded topic models capable of dealing with the homography of words in arbitrary documents.

\section{Topic Models}
\label{sec:background}

Here, we review topic models on which our method is based: LDA, ETM, D-ETM.
In the following, we consider a collection of $D$ documents, where the vocabulary contains $V$ distinct terms.
Let $w_{dn} \in \{1, \dots, V\}$ denote the $n$-th word in the $d$-th document.

\subsection{Latent Dirichlet Allocation (LDA)}

LDA is a probabilistic generative model of documents~\cite{10.5555/944919.944937}.
It posits $K$ topics, and the distribution over the vocabulary for each topic $k$ is represented $\bm{\beta}_k \in \mathbb{R}^V$.
It assumes each document comes from a mixture of topics, where the topics are shared across the given documents and the mixture proportions are unique for each document.
Specifically, LDA considers a vector of topic proportions $\bm{\theta}_d \in \mathbb{R}^K$ for each document $d$; each element $\theta_{dk}$ expresses how prevalent the $k$-th topic is in the document $d$.
In the generative process of LDA, each word is assigned to topic $k$ with the probability $\theta_{dk}$, and the word is then drawn from the distribution $\bm{\beta}_k$.
The generative process for each document is as follows:
\begin{enumerate}
    \item Draw topic proportion: $\bm{\theta}_d \sim \text{Dirichlet}(\bm{\eta}_{\theta})$
    \item For each word $n$ in $d$:
    \begin{enumerate}
        \item Draw topic assignment: $\bm{z}_{dn} \sim \text{Cat}(\bm{\theta}_d)$
        \item Draw word: $w_{dn} \sim \text{Cat}(\bm{\beta}_{\bm{z}_{dn}})$.
    \end{enumerate}
\end{enumerate}
Here, $\text{Cat}(\cdot)$ denotes a categorical distribution.
LDA places a Dirichlet prior on the topics, $\bm{\beta}_k \sim \text{Dirichlet}(\bm{\alpha}_{\beta})$.
The two concentration parameters of the Dirichlet distributions, $\bm{\alpha}_{\beta}$ and $\bm{\eta}_{\theta}$, are fixed model hyperparameters.

\subsection{Embedded Topic Model (ETM)}

ETM~\cite{dieng-etal-2020-topic} is a neural topic model powered by word embeddings~\cite{NIPS2013_9aa42b31} and a neural network.
Here, let $\bm{\rho}$ be an $L \times V$ matrix, which contains $L$-dimensional embeddings of the words in the vocabulary.
Each column $\bm{\rho}_v \in \mathbb{R}^L$ corresponds to the embedding of the $v$-th term.
ETM uses this embedding matrix $\bm{\rho}$ to define the word distribution of each topic, 
$\bm{\beta}_k = \textrm{softmax}(\bm{\rho}^\top \bm{\alpha}_k)$.
$\bm{\alpha}_k$ is an embedding representation of the $k$-th topic in the semantic space of words, called topic embedding.
The generative process of ETM is analogous to LDA as follows:
\begin{enumerate}
    \item Draw topic proportion: $\bm{\theta}_d \sim \mathcal{LN}(\bm{0}, \bm{I})$
    \item For each word $n$ in $d$:
    \begin{enumerate}
        \item Draw topic assignment: $\bm{z}_{dn} \sim \text{Cat}(\bm{\theta}_d)$
        \item Draw word: $w_{dn} \sim \text{Cat}(\bm{\beta}_{\bm{z}_{dn}})$.
    \end{enumerate}
\end{enumerate}
Here, $\mathcal{LN}(\cdot, \cdot)$ denotes a logistic-normal distribution~\cite{10.1093/biomet/67.2.261}.
The intuition behind ETM is that the embedding representations of semantically related words are similar to each other, they will interact with the topic
embeddings $\bm{\alpha}_k$ similarly, and then they will be assigned to similar topics.

\begin{figure*}[t]
    \begin{minipage}[b]{\textwidth}
    \begin{center}
        \vspace{3mm}
        \includegraphics[width=\textwidth]{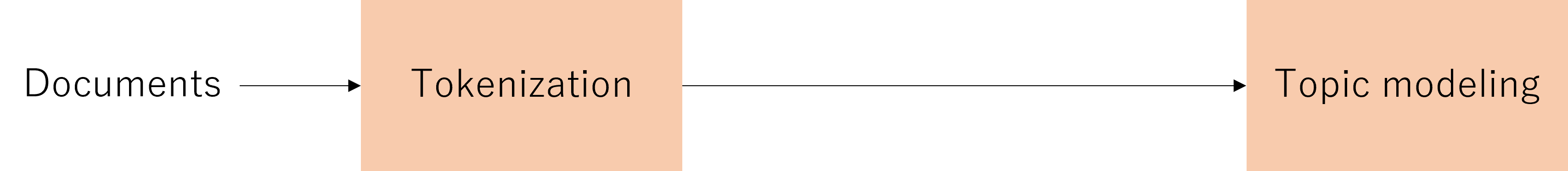}\\
        \vspace{1mm}
        (b) Conventional topic models
    \end{center}
    \end{minipage}
    \begin{minipage}[b]{\textwidth}
    \begin{center}
        \vspace{3mm}
        \includegraphics[width=\textwidth]{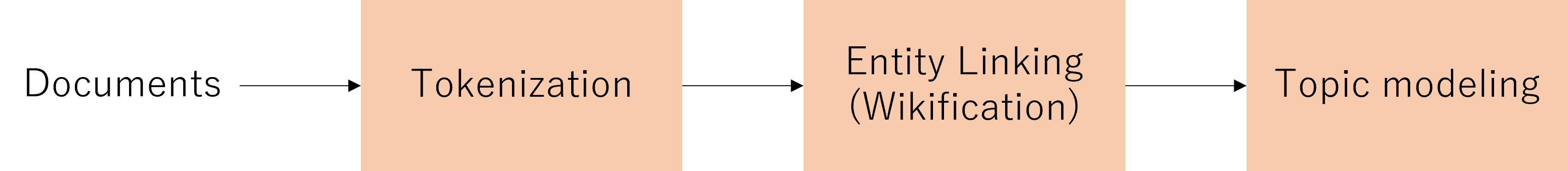}\\
        \vspace{1mm}
        (b) Our proposed topic model
    \end{center}
    \end{minipage}
    \caption{Processing flows of conventional topic models and our proposed topic model.}
    \label{fig:processing_flow}
\end{figure*}

\subsection{Dynamic Embedded Topic Model (D-ETM)}

D-ETM~\cite{dieng2019dynamicembeddedtopicmodel} analyzes time-series documents by introducing Markov chains to the topic embeddings $\bm{\alpha}_k$ and the topic proportion mean.
As in ETM, D-ETM considers an embedding matrix $\bm{\rho} \in \mathbb{R}^{L \times V}$, such that each column $\bm{\rho}_v \in \mathbb{R}^L$ corresponds to the embedding of the $v$-th term.
D-ETM posits an topic embedding $\bm{\alpha}_k^{(t)} \in \mathbb{R}^L$ for each topic $k$ at a time stamp $t \in \{1, \dots, T\}$.
This means D-ETM represents each topic with a time-varying vector.
Then, the word distribution for the $k$-th topic in the time step $t$ is defined by $\bm{\beta}_k^{(t)} = \textrm{softmax}(\bm{\rho}^\top \bm{\alpha}_k^{(t)})$.
Here, the generative process of D-ETM for documents is described as follows:
\begin{enumerate}
    \item For time step $t=0$:
    \begin{enumerate}
        \item Draw initial topic embedding: \\
        $\bm{\alpha}_k^{(0)} \sim \mathcal{N}(\bm{0}, I)$ for $k \in \{1, \ldots, K\}$ 
        \item Draw initial topic proportion mean: \\
        $\bm{\eta}_0 \sim \mathcal{N}(\bm{0}, I)$
    \end{enumerate}
    \item For each time step $t \in \{1, \ldots, T\}$:
    \begin{enumerate}
          \item Draw topic embedding: \\
        $\bm{\alpha}_k^{(t)} \sim \mathcal{N}(\bm{\alpha}_k^{(t-1)}, \sigma^2 I)$ \\
        for $k \in \{1, \ldots, K\}$
          \item Draw topic proportion mean: \\
        $\bm{\eta}_t \sim \mathcal{N}(\bm{\eta}_{t-1}, \delta^2 I)$
    \end{enumerate}
    \item For each document $d \in \{1, \ldots, D\}$:
    \begin{enumerate}
        \item Draw topic proportion: \\
        $\bm{\theta}_d \sim \mathcal{LN}(\bm{\eta}_{t_d}, \gamma^2 I)$
        \item For each word $n$ in $d$:
        \begin{enumerate}
            \item Draw topic assignment: \\
            $\bm{z}_{dn} \sim \text{Cat}(\bm{\theta}_d)$
            \item Draw word: \\
            $w_{dn} \sim \text{Cat}(\bm{\beta}_{\bm{z}_{dn}}^{(t_d)})$,
        \end{enumerate}
    \end{enumerate}
\end{enumerate}
where $\mathcal{N}(\cdot, \cdot)$ denotes a normal distribution distribution.
$\sigma$, $\delta$, and $\gamma$ are model hyperparameters, each of which controls the variance of the corresponding normal distribution.
$t_d$ denotes the time stamp of the document $d$.
Step 2(a) encourages smooth variations of the topic embeddings, and Step 2(b) describes time-varying priors over the topic proportions $\bm{\theta}_d$.

In this study, we incorporate entity knowledge into ETM or D-ETM by utilizing not only word embeddings but also entity embeddings, which enables topic models to be aware of named entities.
To the best of our knowledge, our study is the first attempt to apply entity embeddings to embedded topic models.
In the next section, we will explain how we introduce entity embeddings into embedded topic models.

\section{Proposed Method}
\label{sec:proposed}

In this study, we propose a method of incorporating word disambiguation results into a neural topic model.
We depict the processing flows of conventional topic models and our proposed method in Figure~\ref{fig:processing_flow}.
In previous embedded topic models such as ETM and D-ETM, given documents are first tokenized, and then the word embedding matrix $\bm{\rho}$ is built by tiling the pretrained word embeddings such as skip-gram~\cite{NIPS2013_9aa42b31} corresponding to tokenized words.
On the other hand, we incorporate entity information extracted by entity linking (EL) into the word embedding matrix $\bm{\rho}$ of an ETM/D-ETM.
We explain the details of our method below.

\begin{figure*}[t]
    \begin{center}
       \vspace{3mm}
        \includegraphics[width=\textwidth]{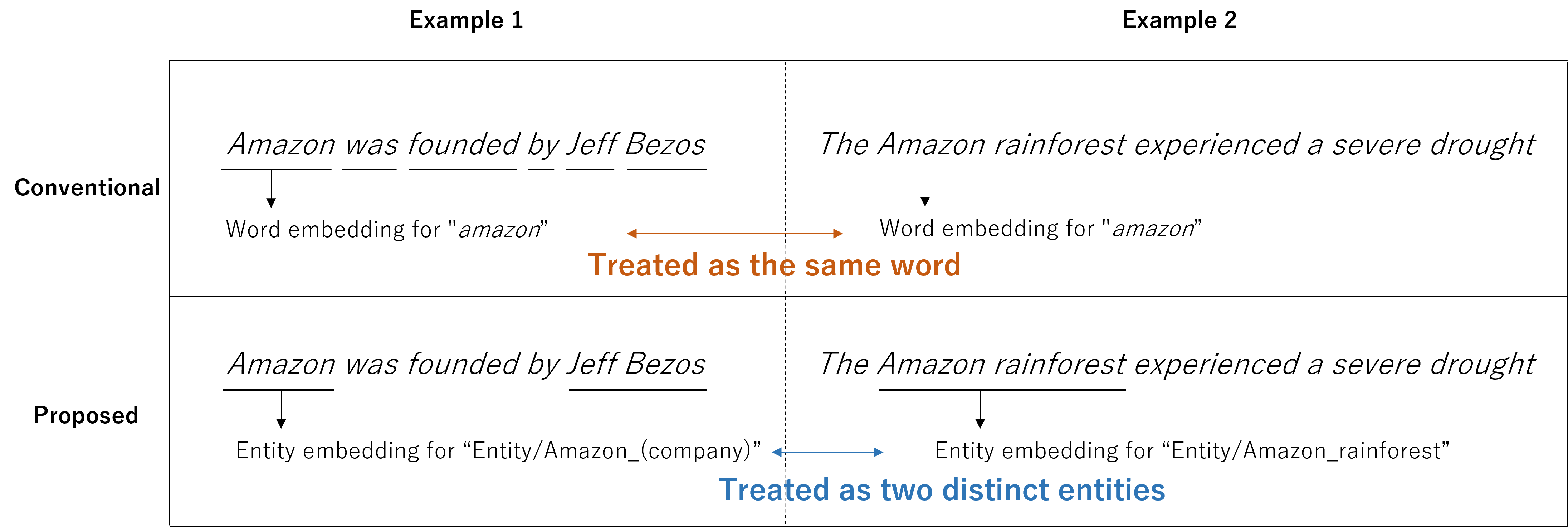}\\
    \end{center}
    \caption{Difference between conventional embedded topic models and our proposed topic model.}
    \label{fig:processing_detail}
\end{figure*}

\subsection{Incorporation of Entity Linking}
\label{sec:entity_linking}

Here, we explain a way of building the embedding matrix $\bm{\rho}$ based on EL results.
EL is a task that assigns a unique identity to an entity mention in text.
In this study, we use an entity embedding instead of a word embedding if an entity linker identifies a phrase in a document as an entry in a knowledge base (KB) as depicted in Figure~\ref{fig:processing_detail}.
Specifically, we utilize entity embedding trained with the Wikipedia2Vec toolkit~\cite{yamada-etal-2020-wikipedia2vec}.
The Wikipedia2Vec toolkit can learn the embeddings of both words and entities by using  Wikipedia's text and hyperlinks.
We can incorporate distributed representations of not only words but also entities into neural models with them.
For example, if a word ``amazon'' is identified as a KB entry ``Amazon (company)'' in a document, we adopt the entity embedding corresponding to ``Amazon (company)''.
If ``\textit{amazon}'' is identified as a KB entry ``Amazon rainforest'' in another document (or another place of the same document), we use the entity embedding for ``Amazon rainforest''.
If ``\textit{amazon}'' is not identified to any KB entry, we adopt the word embedding corresponding to ``\textit{amazon}''.
Thus we deal with the entity ``Amazon (company)'', the entity ``Amazon rainforest'', and the word ``\textit{amazon}'' as distinct items.
Through the above procedure, we can incorporate EL results into a neural topic model and make it aware of named entities.
In the next section, we will evaluate the performance of our proposed method.

\section{Experiments}
\label{sec:experiments}

In this section, we conduct two experiments.
First, we evaluate our method on our original dataset, which requires a topic model to be aware of named entities.
Our first experiment aims to verify that our method is effective in a case where word disambiguation is important.
Next, we evaluate our method on the AIDA-CoNLL dataset~\cite{hoffart-etal-2011-robust}.
The AIDA-CoNLL dataset provides manual entity annotations.
In this second experiment, we aim to assess 1) whether our method of incorporating entity information does not harm the performance of topic models even in a case where word disambiguation is not necessarily required and 2) how largely the off-the-shelve entity linker used in our pipeline deteriorates the performance in comparison with the use of the gold entity annotations.

\begin{table*}[t]
    \centering
    \small
    \begin{tabular}{l|c|c}
        Method & ``\textit{apple}'' & ``\textit{amazon}'' \\
        \hline
        \hline
        ETM & 5753.3 $\pm$ 227.2 & 5086.7 $\pm$ 304.8 \\
        ETM+EL & 5228.9 $\pm$ 730.9 & 6412.4 $\pm$ 731.3 \\
        \hline
        DSNTM~\cite{miyamoto-etal-2023-dynamic} & 4597.9 $\pm$ 270.0 & 4587.6 $\pm$ 349.0 \\
        DSNTM+EL & \textbf{3578.6} $\pm$ 141.4 & 4038.7 $\pm$ \ 65.9 \\
    \end{tabular}
    \caption{Results for perplexity with 95\% confidence interval (CI) on our \textit{New York Times} dataset. The lower, the better. EL means entity linking.}
    \label{table:ppl_all_nyt}
\end{table*}

\begin{figure*}[t]
    \begin{minipage}[b]{\textwidth}
    \begin{center}
        \vspace{3mm}
        \includegraphics[width=\textwidth]{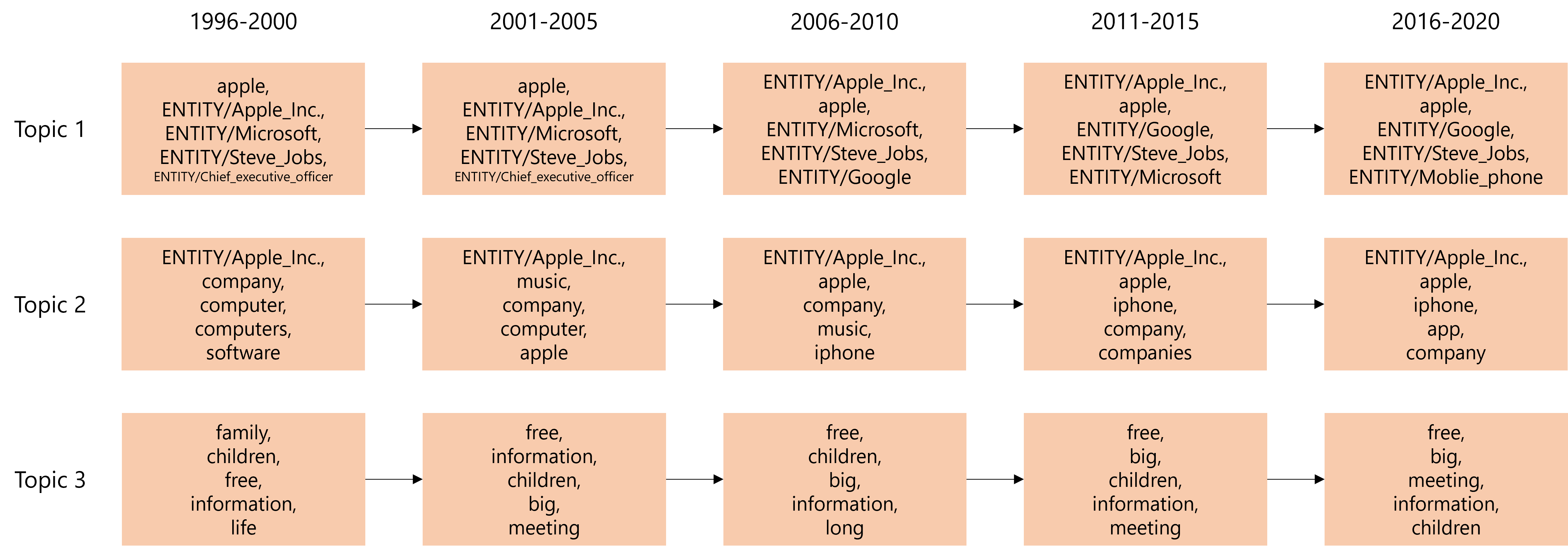}\\
        \vspace{1mm}
        (a) DSNTM+EL
    \end{center}
    \end{minipage}
    \begin{minipage}[b]{\textwidth}
    \begin{center}
        \vspace{3mm}
        \includegraphics[width=\textwidth]{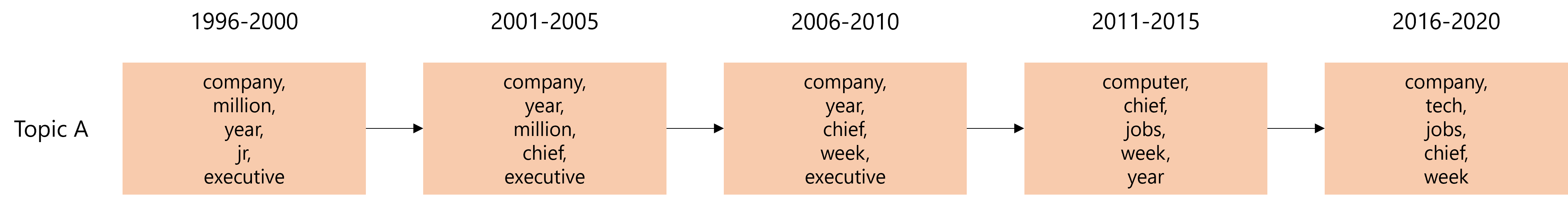}\\
        \vspace{1mm}
        (b) DSNTM
    \end{center}
    \end{minipage}
    \caption{Examples of topic transition. We present the top five most frequent terms in each topic.}
    \label{fig:topic_transition}
\end{figure*}

\subsection{Fine-Grained Topic Modeling}

\subsubsection{Experimental Setup}
\label{sec:setup}

\noindent\textbf{Dataset.}
In this experiment, we use archive news articles of \textit{New York Times}\footnote{\url{https://developer.nytimes.com}}.
We extract two subsets of articles published between the years 1996 and 2020: 1) a collection of 6,651 documents that include the word ``\textit{apple}'' and 2) a collection of 3,070 documents that include ``\textit{amazon}''.
We regard each of the two collections as a single dataset and assess if our proposed method can train a more generalizable topic model by disambiguating homographic words, ``\textit{apple}'' and ``\textit{amazon}''.
We randomly split each collection into 3:1:1 for training, validation, and test sets.
Following \citet{miyamoto-etal-2023-dynamic}, we filter out words that appear in 70\% or more of documents and words included in a predefined stop-word list before building an embedding matrix $\bm{\rho}$.
We group documents published within five consecutive years into a single time step.
For example, news articles published between 1996 and 2000 are grouped.

\noindent\textbf{Compared Models.}
We use \textbf{ETM}~\cite{dieng-etal-2020-topic} and \textbf{DSNTM}~\cite{miyamoto-etal-2023-dynamic} (one implementation of D-ETM~\cite{dieng2019dynamicembeddedtopicmodel}) as baseline models, where only tokenization is applied to documents.
ETM is not a dynamic topic model and does not consider time stamp information, whereas DSNTM is a dynamic topic model and can capture the chronological transition of topics.
We assess if our method is effective in each model.
We compare \textbf{ETM+EL} and \textbf{DSNTM+EL} (where we use entity embeddings for entities identified by an entity linker) with their corresponding baselines to see if our proposed method is effective.

\noindent\textbf{Implementation Details.}
We set the number of topics $K=10$ for all models.
The variances of the prior distributions are set $\delta^2 = \sigma^2 = 0.005$ and $\gamma^2 = 1$.
We use 500-dimensional word/entity embeddings (window size: 10)\footnote{\url{http://wikipedia2vec.s3.amazonaws.com/models/en/2018-04-20/enwiki_20180420_win10_500d.txt.bz2}} pretrained with the Wikipedia2Vec toolkit~\cite{yamada-etal-2020-wikipedia2vec}\footnote{\url{https://wikipedia2vec.github.io/wikipedia2vec/}}. 
Regarding other hyperparameters, we follow the official implementation of DSNTM\footnote{\url{https://github.com/miyamotononno/DSNTM}}.
For the preprocessing of documents, we utilize the tokenizer and entity linker implemented in the Stanford CoreNLP toolkit~\cite{manning-etal-2014-stanford}.\footnote{Although more accurate entity linkers~\cite{shavarani-sarkar-2023-spel,wang-etal-2024-entity} are publicly available, we choose the one implemented in the Stanford CoreNLP due to the limitation of computing resources.}
We call these CoreNLP analyzers through the Stanza library~\cite{qi-etal-2020-stanza}\footnote{\url{https://github.com/stanfordnlp/stanza}}.

\subsubsection{Quantitative Evaluation}
\label{sec:quantitative}

We use perplexity~\cite{10.5555/1036843.1036902} to evaluate the generalizability of a topic model.
Although there is a discussion on how to properly evaluate topic models~\cite{NIPS2009_f92586a2,NEURIPS2021_0f83556a}, perplexity is still a widely-used objective metric~\cite{hida-etal-2018-dynamic,miyamoto-etal-2023-dynamic}.
It measures the ability to predict words in unseen documents.
In training, we apply early stopping based on the performance of a validation set.
We train each model eight times with different random seeds and report the average performance and its 95\% confidence interval on a test set.

The results are shown in Table~\ref{table:ppl_all_nyt}.
We can find two tendencies in the results.
The first one is that EL tends to improve the performance except for ETM on the ``\textit{amazon}'' dataset.
In particular, DSNTM+EL achieves lower perplexity than DSNTM.
This demonstrates that word disambiguation by EL is effective in analyzing a collection of documents with a topic model.
We will discuss the reason why our method does not work well with ETM on the ``\textit{amazon}'' dataset in a later section.
The second tendency is that DSNTM+EL performs better than ETM+EL.
This means that modeling a temporal change of topics is effective even when EL is combined.

\begin{table*}[t]
    \centering
    \small
    \begin{tabular}{ll|c}
        Word/Entity 1 & Word/Entity 2 & Cosine similarity \\
        \hline
        \hline
        ENTITY/Apple\_Inc. & \textit{apple} & 0.67 \\
        ENTITY/Apple\_Inc. & ENTITY/Steve\_Jobs & 0.59 \\
        ENTITY/Apple\_Inc. & \textit{steve} & 0.27 \\
        ENTITY/Apple\_Inc. & \textit{jobs} & 0.28 \\
        \textit{apple} & ENTITY/Steve\_Jobs & 0.52 \\
        \textit{apple} & \textit{steve} & 0.30 \\
        \textit{apple} & \textit{jobs} & 0.30 \\
    \end{tabular}
    \caption{Word similarities of two words/entities on Wikipedia2Vec~\cite{yamada-etal-2020-wikipedia2vec}.}
    \label{table:cos_sim}
\end{table*}

\subsubsection{Qualitative Analysis}
\label{sec:qualitative}

\noindent\textbf{Visualization of Topic Transition.}
We present an overview of the topic transition process extracted by a trained DSNTM+EL model on the ``\textit{apple}'' dataset in Figure~\ref{fig:topic_transition}(a).
The topics in the first, second, and third rows (Topics~1, 2, and 3) represent business/management, products/services, and \textit{New York City}, respectively.
When we look into Topic~1, the word ``\textit{apple}'', the entity ``ENTITY/Apple\_Inc.'', and the entity ``ENTITY/Steve\_Jobs'' are frequently used constantly between 1996 and 2020, whereas the entity ``ENTITY/Google'' emerges after 2006.
This is reasonable because Google was founded in 1998 and went public via an initial public offering (IPO) in 2004.
Google was never mentioned before 1998 and not often before 2004.
This demonstrates that DSNTM+EL successfully finds the transition of frequent terms in each topic and that we can easily understand the trends of topics by visualization.
This is true for ``\textit{iphone}'' (released in 2007) in Topic 2 as well.
Regarding Topic~3, one might think this topic has nothing to do with the word ``\textit{apple}'' at a glance, but this topic is related to \textit{New York City}.
\textit{New York City} sometimes is called its nickname, ``\textit{Big Apple}''.
This topic consists of articles about \textit{New York City}, especially entertainment such as \textit{Big Apple Circus} and \textit{Big Apple Chorus}.
Then, the word ``\textit{big}'' is listed as a frequent term.\footnote{Ideally, entity linkers should recognize those entities correctly, but the entity linker used in our pipeline is not so accurate. As a result, the word ``\textit{big}'' is listed.}

We also show the transition of a topic (Topic A) extracted by a trained DSNTM model in Figure~\ref{fig:topic_transition}(b).
According to the frequent terms, Topic A is similar to Topic 1 in Figure~\ref{fig:topic_transition}(a).
This means that a conventional topic model can analyze documents in a similar way.
However, our method involving entity linking into its preprocessing comes with higher interpretability as frequent terms are expressed with not only words but also entities.
The word ``\textit{jobs}'' in Topic A means \textit{Steve Jobs} in almost all cases, but DSNTM+EL shows that Topic 1 is related to \textit{Steve Jobs} in a much easier-to-understand manner.
This high interpretability is another advantage of our proposed method in addition to lower perplexities.

\noindent\textbf{Influence of Entity Embedding.}
We investigate why entity linking (EL) boosts the performance of neural topic models.
Some words have multiple meanings, whereas previous topic models deal with such words without being aware of meanings, considering only their spelling.
In such an approach, a topic model can take into consideration neither who ``\textit{steve}'' is nor whether ``\textit{jobs}'' is a person's name or a common noun.
In our proposed method, we try to disambiguate words, and use entity embedding trained with the Wikipedia2Vec toolkit~\cite{yamada-etal-2020-wikipedia2vec} instead of conventional word embedding if a word is linked to a KB entry.

Here, let us show some properties of the entity embedding used.
We show the cosine similarities between some words/entities in Table~\ref{table:cos_sim}.
As shown, ``ENTITY/Steve\_Jobs'' is much closer to ``ENTITY/Apple\_Inc.'' than the words ``\textit{steve}'' and ``\textit{jobs}''.
This is because the word ``\textit{jobs}'' can be a noun word (the plural form of ``\textit{job}''), and even ``\textit{steve}'' can be the name of another person.
Then, their embedding vectors are trained in various contexts.
On the other hand, ``ENTITY/Steve\_Jobs'' tends to appear in articles relevant to \textit{Apple Inc.}, and then its entity embedding is trained in a narrow range of contexts.
As a result, the entity embedding of ``ENTITY/Steve\_Jobs'' has a large similarity to the entity embedding of ``ENTITY/Apple\_Inc.'', while the word embeddings of ``\textit{steve}'' and ``\textit{jobs}'' go far from the entity embedding of ``ENTITY/Apple\_Inc.''.

In ETM and DSNTM, a topic embedding $\bm{\alpha}_k^{(t)}$ is multiplied with a static word/entity embedding matrix $\bm{\rho}$ to estimate a distribution of terms, $w_{dn} \sim \text{Cat}(\textrm{softmax}(\bm{\rho}^\top \bm{\alpha}_{z_{dn}}^{(t_d)}))$ (See Section~\ref{sec:background}).
This means that, if word/entity embedding vectors cluster based on their used context, topic embedding can be easily trained.
Actually, entity embedding has such a property as we explained in the previous paragraph.
Thus, entity embedding can help neural topic models extract topics from documents.

\begin{table*}[t]
    \centering
    \small
    \begin{tabular}{l|c|c}
        Method & Tokenization \& entity linking & Perplexity \\
        \hline
        \hline
        ETM & Gold annotation & 5380.5 $\pm$ 246.2 \\
        ETM+EL (ours) & Gold annotation & \textbf{5010.1} $\pm$ 448.8 \\
        \hline
        ETM & Stanford CoreNLP & 5404.9 $\pm$ 225.0 \\
        ETM+EL (ours) & Stanford CoreNLP & 6558.1 $\pm$ 979.4 \\
    \end{tabular}
    \caption{Results for perplexity with 95\% confidence interval (CI) on the AIDA-CoNLL dataset~\cite{hoffart-etal-2011-robust}. The lower, the better. EL means entity linking.}
    \label{table:ppl_all_aida}
\end{table*}

\noindent\textbf{Dependency on Entity Linking.}
In contrast to our aim, entity linking (EL) does not boost the performance of ETM on the ``\textit{amazon}'' dataset, different from the ``\textit{apple}'' dataset.
We find that the accuracy of entity linking is not so good on the ``\textit{amazon}'' dataset and that the entity linker fails to assign entity mentions to correct KB entries.
Our proposed method is a pipeline of 1) preprocessing with an entity linker and 2) neural topic modeling.
If the preprocessing is not accurate, the successive topic modeling will naturally be affected.
We hypothesize that the latest, more accurate entity linkers~\cite{shavarani-sarkar-2023-spel,wang-etal-2024-entity} can boost the performance of neural topic models more.
To verify our hypothesis, we will conduct an experiment on a dataset that contains manual entity annotations in the next section.

\subsection{Coarse-Grained Topic Modeling}

In this section, we evaluate our method on a dataset accompanied with gold entity annotations, to assess 1) whether our method of incorporating entity information does not harm the performance of topic models even in a case where word disambiguation is not necessarily required and 2) how largely the off-the-shelf entity linker used in our pipeline deteriorates the performance in comparison with the use of the gold entity annotations.

\subsubsection{Experimental Setup}

\noindent\textbf{Dataset.}
In this experiment, we use the AIDA-CoNLL dataset~\cite{hoffart-etal-2011-robust}\footnote{\url{https://www.mpi-inf.mpg.de/departments/databases-and-information-systems/research/ambiverse-nlu/aida}}.
This dataset contains manual Wikipedia annotations for the 1,393 Reuters news stories originally published for the CoNLL-2003 Named Entity Recognition Shared Task~\cite{tjong-kim-sang-de-meulder-2003-introduction}.
The number of Wikipedia annotations is 27,817.
The dataset consists of \texttt{train}, \texttt{testa}, and \texttt{testb} splits, which contain 946, 216, and 231 documents, respectively.
We utilize the three splits as training, validation, and test sets.
As in our previous experiment, we filter out words that appear in 70\% or more of documents and words included in the predefined stop-word list before building an embedding matrix $\bm{\rho}$.
In contrast to the \textit{New York Times} dataset used in the previous experiment, which is created by collecting news articles that include a specific word such as ``\textit{apple}'', the AIDA-CoNLL dataset was made without such an intention.
It should include much less ambiguous words.

\noindent\textbf{Compared Models.}
We use \textbf{ETM}~\cite{dieng-etal-2020-topic} as a baseline model.
As the AIDA-CoNLL dataset provides gold annotations of entity linking (including tokenization), we can assess the influence of the off-the-shelf tokenizer and entity linker on the performance of our entire pipeline by comparing results from using gold annotations and results from using annotations by the tokenizer and entity linker.
Therefore, we evaluate the following four models. 1) \textbf{ETM} that utilizes the gold annotations, 2) \textbf{ETM+EL} that uses the gold annotations, 3) \textbf{ETM} that utilizes annotations provided by Stanford CoreNLP, and 4) \textbf{ETM+EL} that uses annotations given by Stanford CoreNLP.
Since the AIDA-CoNLL dataset does not include time stamp information, we do not adopt a dynamic topic model in this experiment.

\noindent\textbf{Implementation Details.}
In this experiment, we use 300-dimensional word/entity embeddings (window size: 10)\footnote{\url{http://wikipedia2vec.s3.amazonaws.com/models/en/2018-04-20/enwiki_20180420_win10_300d.txt.bz2}} because we encountered training instability with 500-dimensional word/entity embeddings.
Regarding all other hyperparameters and implementations, we follow the previous experiment.

\subsubsection{Results}
The results are shown in Table~\ref{table:ppl_all_aida}.
First, we can see that when the gold annotations are provided, entity linking improves the performance of ETM, even though the used AIDA-CoNLL dataset does not include as many homographic words as our \textit{New York Times} dataset used in the previous experiment.
This demonstrates that our method is potentially generalizable and can perform well on various data.
Second, we observe that using information annotated by the Stanford CoreNLP entity linker deteriorates the performance.
As the knowledge base supported by the entity linker is not identical to that used for the annotations in the AIDA-CoNLL dataset, the accuracy of the entity linker can not be calculated so easily.
However, we can attribute the performance gap between the two cases, 1) gold annotations and 2) the CoreNLP entity linker, to the accuracy of the entity linker.
We believe that the latest, more accurate entity linkers~\cite{shavarani-sarkar-2023-spel,wang-etal-2024-entity} can boost the performance of neural topic models.

\section{Conclusion}
\label{sec:conc}

In this study, we proposed a method of analyzing a collection of documents after disambiguating homographic words.
We incorporated entity information extracted by entity linking into neural topic models.
Our experimental results demonstrated that entity linking improves the generalizability of topic models by disambiguating words such as ``\textit{apple}'' and ``\textit{amazon}''.
In addition, our method offers higher interpretability as frequent terms in each topic are represented with not only words but also entities.

\section*{Limitations}
Our models heavily rely on word/entity embedding as with other neural topic models.
If the word/entity embedding contains some bias, our models will be affected by the bias.

Besides, topic models, including our models, sometimes infer incorrect information about topics, such as the frequent terms appearing in topics, the topic proportion in each document, and the dependencies among topics.
There would be the potential risk of inducing misunderstandings among users.

\section*{Ethics Statement}
Our study complies with the ACL Ethics Policy.
We used \textit{PyTorch} (BSD-style license), \textit{New York Times} articles\footnote{\url{https://developer.nytimes.com/terms}}, the AIDA-CoNLL dataset (Creative Commons Attribution 3.0 license).
Our study was conducted under their licenses and terms.

\section*{Acknowledgments}
We thank anonymous reviewers for helpful feedback on our draft.


\bibliography{custom}

\begin{thebibliography}{42}
\providecommand{\natexlab}[1]{#1}

\bibitem[{Atchison and Shen(1980)}]{10.1093/biomet/67.2.261}
J.~Atchison and S.M. Shen. 1980.
\newblock \href {https://doi.org/10.1093/biomet/67.2.261} {{Logistic-normal distributions:Some properties and uses}}.
\newblock \emph{Biometrika}, 67(2):261--272.

\bibitem[{Batmanghelich et~al.(2016)Batmanghelich, Saeedi, Narasimhan, and Gershman}]{batmanghelich-etal-2016-nonparametric}
Kayhan Batmanghelich, Ardavan Saeedi, Karthik Narasimhan, and Sam Gershman. 2016.
\newblock \href {https://doi.org/10.18653/v1/P16-2087} {Nonparametric spherical topic modeling with word embeddings}.
\newblock In \emph{Proceedings of the 54th Annual Meeting of the Association for Computational Linguistics (Volume 2: Short Papers)}, pages 537--542, Berlin, Germany. Association for Computational Linguistics.

\bibitem[{Blei and Lafferty(2006)}]{10.1145/1143844.1143859}
David~M. Blei and John~D. Lafferty. 2006.
\newblock \href {https://doi.org/10.1145/1143844.1143859} {Dynamic topic models}.
\newblock In \emph{Proceedings of the 23rd International Conference on Machine Learning}, ICML '06, page 113–120, New York, NY, USA. Association for Computing Machinery.

\bibitem[{Blei et~al.(2003)Blei, Ng, and Jordan}]{10.5555/944919.944937}
David~M. Blei, Andrew~Y. Ng, and Michael~I. Jordan. 2003.
\newblock Latent {D}irichlet allocation.
\newblock \emph{J. Mach. Learn. Res.}, 3(null):993–1022.

\bibitem[{Bordes et~al.(2011)Bordes, Weston, Collobert, and Bengio}]{Bordes_Weston_Collobert_Bengio_2011}
Antoine Bordes, Jason Weston, Ronan Collobert, and Yoshua Bengio. 2011.
\newblock \href {https://doi.org/10.1609/aaai.v25i1.7917} {Learning structured embeddings of knowledge bases}.
\newblock \emph{Proceedings of the AAAI Conference on Artificial Intelligence}, 25(1):301--306.

\bibitem[{Boyd-Graber and Blei(2009)}]{10.5555/1795114.1795124}
Jordan Boyd-Graber and David~M. Blei. 2009.
\newblock Multilingual topic models for unaligned text.
\newblock In \emph{Proceedings of the Twenty-Fifth Conference on Uncertainty in Artificial Intelligence}, UAI '09, page 75–82, Arlington, Virginia, USA. AUAI Press.

\bibitem[{Card et~al.(2018)Card, Tan, and Smith}]{card-etal-2018-neural}
Dallas Card, Chenhao Tan, and Noah~A. Smith. 2018.
\newblock \href {https://doi.org/10.18653/v1/P18-1189} {Neural models for documents with metadata}.
\newblock In \emph{Proceedings of the 56th Annual Meeting of the Association for Computational Linguistics (Volume 1: Long Papers)}, pages 2031--2040, Melbourne, Australia. Association for Computational Linguistics.

\bibitem[{Chang et~al.(2009)Chang, Gerrish, Wang, Boyd-graber, and Blei}]{NIPS2009_f92586a2}
Jonathan Chang, Sean Gerrish, Chong Wang, Jordan Boyd-graber, and David Blei. 2009.
\newblock \href {https://proceedings.neurips.cc/paper_files/paper/2009/file/f92586a25bb3145facd64ab20fd554ff-Paper.pdf} {Reading tea leaves: How humans interpret topic models}.
\newblock In \emph{Advances in Neural Information Processing Systems}, volume~22. Curran Associates, Inc.

\bibitem[{Cong et~al.(2017)Cong, Chen, Liu, and Zhou}]{pmlr-v70-cong17a}
Yulai Cong, Bo~Chen, Hongwei Liu, and Mingyuan Zhou. 2017.
\newblock \href {https://proceedings.mlr.press/v70/cong17a.html} {Deep latent {D}irichlet allocation with topic-layer-adaptive stochastic gradient {R}iemannian {MCMC}}.
\newblock In \emph{Proceedings of the 34th International Conference on Machine Learning}, volume~70 of \emph{Proceedings of Machine Learning Research}, pages 864--873. PMLR.

\bibitem[{Cvejoski et~al.(2023)Cvejoski, Sánchez, and Ojeda}]{Cvejoski_Sánchez_Ojeda_2023}
Kostadin Cvejoski, Ramsés~J. Sánchez, and César Ojeda. 2023.
\newblock \href {https://doi.org/10.1609/aaai.v37i11.26496} {Neural dynamic focused topic model}.
\newblock \emph{Proceedings of the AAAI Conference on Artificial Intelligence}, 37(11):12719--12727.

\bibitem[{Das et~al.(2015)Das, Zaheer, and Dyer}]{das-etal-2015-gaussian}
Rajarshi Das, Manzil Zaheer, and Chris Dyer. 2015.
\newblock \href {https://doi.org/10.3115/v1/P15-1077} {{G}aussian {LDA} for topic models with word embeddings}.
\newblock In \emph{Proceedings of the 53rd Annual Meeting of the Association for Computational Linguistics and the 7th International Joint Conference on Natural Language Processing (Volume 1: Long Papers)}, pages 795--804, Beijing, China. Association for Computational Linguistics.

\bibitem[{Dieng et~al.(2019)Dieng, Ruiz, and Blei}]{dieng2019dynamicembeddedtopicmodel}
Adji~B. Dieng, Francisco J.~R. Ruiz, and David~M. Blei. 2019.
\newblock \href {https://arxiv.org/abs/1907.05545} {The dynamic embedded topic model}.
\newblock \emph{Preprint}, arXiv:1907.05545.

\bibitem[{Dieng et~al.(2020)Dieng, Ruiz, and Blei}]{dieng-etal-2020-topic}
Adji~B. Dieng, Francisco J.~R. Ruiz, and David~M. Blei. 2020.
\newblock \href {https://doi.org/10.1162/tacl_a_00325} {Topic modeling in embedding spaces}.
\newblock \emph{Transactions of the Association for Computational Linguistics}, 8:439--453.

\bibitem[{Hao and Paul(2018)}]{hao-paul-2018-learning}
Shudong Hao and Michael~J. Paul. 2018.
\newblock \href {https://aclanthology.org/C18-1220} {Learning multilingual topics from incomparable corpora}.
\newblock In \emph{Proceedings of the 27th International Conference on Computational Linguistics}, pages 2595--2609, Santa Fe, New Mexico, USA. Association for Computational Linguistics.

\bibitem[{Hida et~al.(2018)Hida, Takeishi, Yairi, and Hori}]{hida-etal-2018-dynamic}
Rem Hida, Naoya Takeishi, Takehisa Yairi, and Koichi Hori. 2018.
\newblock \href {https://doi.org/10.18653/v1/P18-2082} {Dynamic and static topic model for analyzing time-series document collections}.
\newblock In \emph{Proceedings of the 56th Annual Meeting of the Association for Computational Linguistics (Volume 2: Short Papers)}, pages 516--520, Melbourne, Australia. Association for Computational Linguistics.

\bibitem[{Hoffart et~al.(2011)Hoffart, Yosef, Bordino, F{\"u}rstenau, Pinkal, Spaniol, Taneva, Thater, and Weikum}]{hoffart-etal-2011-robust}
Johannes Hoffart, Mohamed~Amir Yosef, Ilaria Bordino, Hagen F{\"u}rstenau, Manfred Pinkal, Marc Spaniol, Bilyana Taneva, Stefan Thater, and Gerhard Weikum. 2011.
\newblock \href {https://aclanthology.org/D11-1072} {Robust disambiguation of named entities in text}.
\newblock In \emph{Proceedings of the 2011 Conference on Empirical Methods in Natural Language Processing}, pages 782--792, Edinburgh, Scotland, UK. Association for Computational Linguistics.

\bibitem[{Hoyle et~al.(2021)Hoyle, Goel, Hian-Cheong, Peskov, Boyd-Graber, and Resnik}]{NEURIPS2021_0f83556a}
Alexander Hoyle, Pranav Goel, Andrew Hian-Cheong, Denis Peskov, Jordan Boyd-Graber, and Philip Resnik. 2021.
\newblock \href {https://proceedings.neurips.cc/paper_files/paper/2021/file/0f83556a305d789b1d71815e8ea4f4b0-Paper.pdf} {Is automated topic model evaluation broken? the incoherence of coherence}.
\newblock In \emph{Advances in Neural Information Processing Systems}, volume~34, pages 2018--2033. Curran Associates, Inc.

\bibitem[{Kingma and Welling(2014)}]{pmlr-v32-kingma14}
Diederik Kingma and Max Welling. 2014.
\newblock \href {https://proceedings.mlr.press/v32/kingma14.html} {Efficient gradient-based inference through transformations between {B}ayes nets and neural nets}.
\newblock In \emph{Proceedings of the 31st International Conference on Machine Learning}, volume~32 of \emph{Proceedings of Machine Learning Research}, pages 1782--1790, Bejing, China. PMLR.

\bibitem[{Lin et~al.(2015)Lin, Liu, Sun, Liu, and Zhu}]{Lin_Liu_Sun_Liu_Zhu_2015}
Yankai Lin, Zhiyuan Liu, Maosong Sun, Yang Liu, and Xuan Zhu. 2015.
\newblock \href {https://doi.org/10.1609/aaai.v29i1.9491} {Learning entity and relation embeddings for knowledge graph completion}.
\newblock \emph{Proceedings of the AAAI Conference on Artificial Intelligence}, 29(1).

\bibitem[{Manning et~al.(2014)Manning, Surdeanu, Bauer, Finkel, Bethard, and McClosky}]{manning-etal-2014-stanford}
Christopher Manning, Mihai Surdeanu, John Bauer, Jenny Finkel, Steven Bethard, and David McClosky. 2014.
\newblock \href {https://doi.org/10.3115/v1/P14-5010} {The {S}tanford {C}ore{NLP} natural language processing toolkit}.
\newblock In \emph{Proceedings of 52nd Annual Meeting of the Association for Computational Linguistics: System Demonstrations}, pages 55--60, Baltimore, Maryland. Association for Computational Linguistics.

\bibitem[{Mikolov et~al.(2013)Mikolov, Sutskever, Chen, Corrado, and Dean}]{NIPS2013_9aa42b31}
Tomas Mikolov, Ilya Sutskever, Kai Chen, Greg~S Corrado, and Jeff Dean. 2013.
\newblock \href {https://proceedings.neurips.cc/paper_files/paper/2013/file/9aa42b31882ec039965f3c4923ce901b-Paper.pdf} {Distributed representations of words and phrases and their compositionality}.
\newblock In \emph{Advances in Neural Information Processing Systems}, volume~26. Curran Associates, Inc.

\bibitem[{Miyamoto et~al.(2023)Miyamoto, Isonuma, Takase, Mori, and Sakata}]{miyamoto-etal-2023-dynamic}
Nozomu Miyamoto, Masaru Isonuma, Sho Takase, Junichiro Mori, and Ichiro Sakata. 2023.
\newblock \href {https://doi.org/10.18653/v1/2023.findings-acl.366} {Dynamic structured neural topic model with self-attention mechanism}.
\newblock In \emph{Findings of the Association for Computational Linguistics: ACL 2023}, pages 5916--5930, Toronto, Canada. Association for Computational Linguistics.

\bibitem[{Miz et~al.(2020)Miz, Hanna, Aspert, Ricaud, and Vandergheynst}]{10.1145/3366424.3383567}
Volodymyr Miz, Jo\"{e}lle Hanna, Nicolas Aspert, Benjamin Ricaud, and Pierre Vandergheynst. 2020.
\newblock \href {https://doi.org/10.1145/3366424.3383567} {What is trending on {W}ikipedia? capturing trends and language biases across {W}ikipedia editions}.
\newblock In \emph{Companion Proceedings of the Web Conference 2020}, WWW '20, page 794–801, New York, NY, USA. Association for Computing Machinery.

\bibitem[{Ni et~al.(2009)Ni, Sun, Hu, and Chen}]{10.1145/1526709.1526904}
Xiaochuan Ni, Jian-Tao Sun, Jian Hu, and Zheng Chen. 2009.
\newblock \href {https://doi.org/10.1145/1526709.1526904} {Mining multilingual topics from {W}ikipedia}.
\newblock In \emph{Proceedings of the 18th International Conference on World Wide Web}, WWW '09, page 1155–1156, New York, NY, USA. Association for Computing Machinery.

\bibitem[{Petterson et~al.(2010)Petterson, Buntine, Narayanamurthy, Caetano, and Smola}]{NIPS2010_db85e259}
James Petterson, Wray Buntine, Shravan Narayanamurthy, Tib\'{e}rio Caetano, and Alex Smola. 2010.
\newblock \href {https://proceedings.neurips.cc/paper_files/paper/2010/file/db85e2590b6109813dafa101ceb2faeb-Paper.pdf} {Word features for latent {D}irichlet allocation}.
\newblock In \emph{Advances in Neural Information Processing Systems}, volume~23. Curran Associates, Inc.

\bibitem[{Piccardi and West(2021)}]{10.1145/3442381.3449805}
Tiziano Piccardi and Robert West. 2021.
\newblock \href {https://doi.org/10.1145/3442381.3449805} {Crosslingual topic modeling with {W}iki{PDA}}.
\newblock In \emph{Proceedings of the Web Conference 2021}, WWW '21, page 3032–3041, New York, NY, USA. Association for Computing Machinery.

\bibitem[{Qi et~al.(2020)Qi, Zhang, Zhang, Bolton, and Manning}]{qi-etal-2020-stanza}
Peng Qi, Yuhao Zhang, Yuhui Zhang, Jason Bolton, and Christopher~D. Manning. 2020.
\newblock \href {https://doi.org/10.18653/v1/2020.acl-demos.14} {{S}tanza: A python natural language processing toolkit for many human languages}.
\newblock In \emph{Proceedings of the 58th Annual Meeting of the Association for Computational Linguistics: System Demonstrations}, pages 101--108, Online. Association for Computational Linguistics.

\bibitem[{Rosen-Zvi et~al.(2004)Rosen-Zvi, Griffiths, Steyvers, and Smyth}]{10.5555/1036843.1036902}
Michal Rosen-Zvi, Thomas Griffiths, Mark Steyvers, and Padhraic Smyth. 2004.
\newblock The author-topic model for authors and documents.
\newblock In \emph{Proceedings of the 20th Conference on Uncertainty in Artificial Intelligence}, UAI '04, page 487–494, Arlington, Virginia, USA. AUAI Press.

\bibitem[{Shavarani and Sarkar(2023)}]{shavarani-sarkar-2023-spel}
Hassan Shavarani and Anoop Sarkar. 2023.
\newblock \href {https://doi.org/10.18653/v1/2023.emnlp-main.686} {{S}p{EL}: Structured prediction for entity linking}.
\newblock In \emph{Proceedings of the 2023 Conference on Empirical Methods in Natural Language Processing}, pages 11123--11137, Singapore. Association for Computational Linguistics.

\bibitem[{Socher et~al.(2013)Socher, Chen, Manning, and Ng}]{NIPS2013_b337e84d}
Richard Socher, Danqi Chen, Christopher~D Manning, and Andrew Ng. 2013.
\newblock \href {https://proceedings.neurips.cc/paper_files/paper/2013/file/b337e84de8752b27eda3a12363109e80-Paper.pdf} {Reasoning with neural tensor networks for knowledge base completion}.
\newblock In \emph{Advances in Neural Information Processing Systems}, volume~26. Curran Associates, Inc.

\bibitem[{Tjong Kim~Sang and De~Meulder(2003)}]{tjong-kim-sang-de-meulder-2003-introduction}
Erik~F. Tjong Kim~Sang and Fien De~Meulder. 2003.
\newblock \href {https://aclanthology.org/W03-0419} {Introduction to the {C}o{NLL}-2003 shared task: Language-independent named entity recognition}.
\newblock In \emph{Proceedings of the Seventh Conference on Natural Language Learning at {HLT}-{NAACL} 2003}, pages 142--147.

\bibitem[{Wang et~al.(2024)Wang, Mousavi, Attia, Pradeep, Potdar, Rush, Minhas, and Li}]{wang-etal-2024-entity}
Junxiong Wang, Ali Mousavi, Omar Attia, Ronak Pradeep, Saloni Potdar, Alexander Rush, Umar~Farooq Minhas, and Yunyao Li. 2024.
\newblock \href {https://doi.org/10.18653/v1/2024.naacl-long.363} {Entity disambiguation via fusion entity decoding}.
\newblock In \emph{Proceedings of the 2024 Conference of the North American Chapter of the Association for Computational Linguistics: Human Language Technologies (Volume 1: Long Papers)}, pages 6524--6536, Mexico City, Mexico. Association for Computational Linguistics.

\bibitem[{Wang et~al.(2014)Wang, Zhang, Feng, and Chen}]{wang-etal-2014-knowledge}
Zhen Wang, Jianwen Zhang, Jianlin Feng, and Zheng Chen. 2014.
\newblock \href {https://doi.org/10.3115/v1/D14-1167} {Knowledge graph and text jointly embedding}.
\newblock In \emph{Proceedings of the 2014 Conference on Empirical Methods in Natural Language Processing ({EMNLP})}, pages 1591--1601, Doha, Qatar. Association for Computational Linguistics.

\bibitem[{Xie et~al.(2015)Xie, Yang, and Xing}]{xie-etal-2015-incorporating}
Pengtao Xie, Diyi Yang, and Eric Xing. 2015.
\newblock \href {https://doi.org/10.3115/v1/N15-1074} {Incorporating word correlation knowledge into topic modeling}.
\newblock In \emph{Proceedings of the 2015 Conference of the North {A}merican Chapter of the Association for Computational Linguistics: Human Language Technologies}, pages 725--734, Denver, Colorado. Association for Computational Linguistics.

\bibitem[{Xun et~al.(2016)Xun, Gopalakrishnan, Ma, Li, Gao, and Zhang}]{7837989}
Guangxu Xun, Vishrawas Gopalakrishnan, Fenglong Ma, Yaliang Li, Jing Gao, and Aidong Zhang. 2016.
\newblock \href {https://doi.org/10.1109/ICDM.2016.0176} {Topic discovery for short texts using word embeddings}.
\newblock In \emph{2016 IEEE 16th International Conference on Data Mining (ICDM)}, pages 1299--1304.

\bibitem[{Xun et~al.(2017)Xun, Li, Zhao, Gao, and Zhang}]{ijcai2017p588}
Guangxu Xun, Yaliang Li, Wayne~Xin Zhao, Jing Gao, and Aidong Zhang. 2017.
\newblock \href {https://doi.org/10.24963/ijcai.2017/588} {A correlated topic model using word embeddings}.
\newblock In \emph{Proceedings of the Twenty-Sixth International Joint Conference on Artificial Intelligence, {IJCAI-17}}, pages 4207--4213.

\bibitem[{Yaghoobzadeh and Sch{\"u}tze(2015)}]{yaghoobzadeh-schutze-2015-corpus}
Yadollah Yaghoobzadeh and Hinrich Sch{\"u}tze. 2015.
\newblock \href {https://doi.org/10.18653/v1/D15-1083} {Corpus-level fine-grained entity typing using contextual information}.
\newblock In \emph{Proceedings of the 2015 Conference on Empirical Methods in Natural Language Processing}, pages 715--725, Lisbon, Portugal. Association for Computational Linguistics.

\bibitem[{Yamada et~al.(2020)Yamada, Asai, Sakuma, Shindo, Takeda, Takefuji, and Matsumoto}]{yamada-etal-2020-wikipedia2vec}
Ikuya Yamada, Akari Asai, Jin Sakuma, Hiroyuki Shindo, Hideaki Takeda, Yoshiyasu Takefuji, and Yuji Matsumoto. 2020.
\newblock \href {https://doi.org/10.18653/v1/2020.emnlp-demos.4} {{W}ikipedia2{V}ec: An efficient toolkit for learning and visualizing the embeddings of words and entities from {W}ikipedia}.
\newblock In \emph{Proceedings of the 2020 Conference on Empirical Methods in Natural Language Processing: System Demonstrations}, pages 23--30, Online. Association for Computational Linguistics.

\bibitem[{Yamada et~al.(2016)Yamada, Shindo, Takeda, and Takefuji}]{yamada-etal-2016-joint}
Ikuya Yamada, Hiroyuki Shindo, Hideaki Takeda, and Yoshiyasu Takefuji. 2016.
\newblock \href {https://doi.org/10.18653/v1/K16-1025} {Joint learning of the embedding of words and entities for named entity disambiguation}.
\newblock In \emph{Proceedings of the 20th {SIGNLL} Conference on Computational Natural Language Learning}, pages 250--259, Berlin, Germany. Association for Computational Linguistics.

\bibitem[{Zhang et~al.(2018)Zhang, Chen, Guo, and Zhou}]{zhang2018whai}
Hao Zhang, Bo~Chen, Dandan Guo, and Mingyuan Zhou. 2018.
\newblock \href {https://openreview.net/forum?id=S1cZsf-RW} {{WHAI}: Weibull hybrid autoencoding inference for deep topic modeling}.
\newblock In \emph{International Conference on Learning Representations}.

\bibitem[{Zhang et~al.(2013)Zhang, Liu, and Zhao}]{10.5555/2540128.2540447}
Tao Zhang, Kang Liu, and Jun Zhao. 2013.
\newblock Cross lingual entity linking with bilingual topic model.
\newblock In \emph{Proceedings of the Twenty-Third International Joint Conference on Artificial Intelligence}, IJCAI '13, page 2218–2224. AAAI Press.

\bibitem[{Zhao et~al.(2017)Zhao, Du, Buntine, and Liu}]{8215536}
He~Zhao, Lan Du, Wray Buntine, and Gang Liu. 2017.
\newblock \href {https://doi.org/10.1109/ICDM.2017.73} {{MetaLDA}: A topic model that efficiently incorporates meta information}.
\newblock In \emph{2017 IEEE International Conference on Data Mining (ICDM)}, pages 635--644.

\end{thebibliography}

\appendix

\end{document}